\documentclass[letterpaper]{article}
\usepackage{uai2019}
\usepackage[margin=1in]{geometry}

\usepackage{booktabs}
\usepackage{url}  
\usepackage{graphicx}  
\usepackage{amsmath,amsfonts,amsthm,amssymb}
\usepackage{setspace}
\usepackage{fancyhdr}
\usepackage{lastpage}
\usepackage[toc,page]{appendix}
\usepackage{extramarks}
\usepackage{chngpage}
\usepackage{soul,color}
\usepackage{xspace}
\usepackage{float,wrapfig}
\usepackage{subcaption}
\usepackage{hyperref}
\usepackage{enumitem}
\usepackage{times}
\usepackage{thmtools}
\usepackage{thm-restate}
\usepackage{hyperref}
\usepackage{cleveref}
\declaretheorem[name=Algorithm,unnumbered=true]{algo*}

\newcommand{\OPT}{\mathsf{OPT}}
\newcommand{\ALG}{\mathsf{ALG}}
\newcommand{\dist}{\delta}

\newcommand{\FirstFit}{{\sc FirstFit}\xspace}
\newcommand{\BestFit}{{\sc BestScore}\xspace}
\newcommand{\RandomBestFit}{{\sc RandomBestScore}\xspace}
\newcommand{\BestFitA}{{\sc BestScore-A}\xspace}
\newcommand{\BestFitR}{{\sc BestScore-R}\xspace}
\newcommand{\RandomBestFitA}{{\sc RandomBestScore-A}\xspace}
\newcommand{\RandomBestFitR}{{\sc RandomBestScore-R}\xspace}

\usepackage{algorithm}
\usepackage[noend]{algpseudocode}
\newtheorem{theorem}{Theorem}

\newtheorem{lemma}{Lemma}

\newenvironment{proof-sketch}{\noindent{\emph{Proof-sketch}}\hspace*{1em}}{\qed\medskip}

\usepackage{color}

\newcommand{\UpdateDist}{{\sc UpdateProbDist}}

\newcommand{\f}{{\mathsf{CST}}}

\title{Dynamic Trip-Vehicle Dispatch with Scheduled and On-Demand Requests}

\author{} 

\author{ {\bf Taoan Huang } \\
Tsinghua University \\
\And\\
{\bf Bohui Fang}  \\
Shanghai Jiaotong University          \\
\And
{\bf Xiaohui Bei}   \\
Nanyang Technological University \\
\And\\
{\bf Fei Fang}\\
Carnegie Mellon University\\
}
\author{ {\bf Taoan Huang$^1$, Bohui Fang$^2$, Xiaohui Bei$^3$, Fei Fang$^4$ } \\
$^1$Tsinghua University, $^2$Shanghai Jiaotong University, \\$^3$Nanyang Technological University, $^4$Carnegie Mellon University \\
}

\begin{document}

\maketitle
\newdimen\origiwspc%
  \newdimen\origiwstr%
  \origiwspc=\fontdimen2\font
  \origiwstr=\fontdimen3\font
  \fontdimen2\font=\origiwspc
  \fontdimen3\font=0.1em
\begin{abstract}
Transportation service providers that dispatch drivers and vehicles to riders start to support both on-demand ride requests posted in real time and rides scheduled in advance, leading to new challenges which, to the best of our knowledge, have not been addressed by existing works. To fill the gap, we design novel trip-vehicle dispatch algorithms to handle both types of requests while taking into account an estimated request distribution of on-demand requests. At the core of the algorithms is the newly proposed Constrained Spatio-Temporal value function (CST-function), which is polynomial-time computable and represents the expected value a vehicle could gain with the constraint that it needs to arrive at a specific location at a given time. Built upon CST-function, we design a randomized best-fit algorithm for scheduled requests and an online planning algorithm for on-demand requests given the scheduled requests as constraints. We evaluate the algorithms through extensive experiments on a real-world dataset of an online ride-hailing platform.
\end{abstract}

\newcommand{\cM}{\mathcal{M}}
\newcommand{\cR}{\mathcal{R}}
\newcommand{\cT}{\mathcal{T}}
\newcommand{\cD}{\mathcal{D}}
\newcommand{\cP}{\mathcal{P}}
\newcommand{\cW}{\mathcal{W}}

\newcommand{\alert}[1]{{\color{red} #1}}

\section{INTRODUCTION}

The growth in location-tracking technology, the popularity of smartphones, and the reduced cost in mobile network communications have led to a revolution in mobility and the prevalent use of on-demand transportation systems, with a tremendous positive societal impact on personal mobility, pollution, and congestion.
Recently, more and more transportation service providers start to support both on-demand ride requests posted in real time and rides scheduled in advance, providing riders with more flexible and reliable service. 
For example, ride-hailing platforms such as Uber match drivers or vehicles (we will use drivers and vehicles interchangeably) and riders in real time upon riders' request and also allow the riders to schedule rides in advance.
Companies like Curb, Shenzhou Zhuanche and ComfortDelGro and Supershuttle transform traditional taxi-hailing, chauffeured car service, and shuttle service to satisfy both types of requests \cite{shuttle,shuttle2}.

The presence of both scheduled requests and on-demand requests leads to new challenges in the task of trip-vehicle dispatch to service providers.
Accepting scheduled requests is a double-edged sword: on one hand, such requests reduce the demand uncertainty and give the service provider more time to prepare and optimize these trips;
On the other hand, certain scheduled requests may lead to waste in time on the way to serve them or prevent the assigned vehicle from serving more valuable on-demand requests. 
Some systems such as Uber simply treat scheduled requests as regular on-demand requests when their pick-up time is due and directly apply an existing dispatch algorithm for on-demand requests. 
Such practice, however, overlooks a fundamental difference between scheduled and on-demand requests: the rider expect to be picked up on time for sure once their scheduled request is accepted, i.e., it is a commitment that the platform \textit{must} fulfill.
Many scheduled requests are for important purposes, such as going to the airport to catch a flight or attending an important meeting. Failing to serve these rides could hurt the credibility of the service provider and its long-term sustainability.
To the best of our knowledge, no existing work can deal with these essential challenges.

In this paper, we fill the gap and provide the first study on trip-vehicle dispatch with both scheduled and on-demand requests. We consider a two-stage decision-making process. In the first stage (Stage 1), the system is presented with a sequence of scheduled requests. The system needs to select which requests to accept and decide how to dispatch vehicles to the accepted requests in an online fashion. In the second stage (Stage 2), the system needs to dispatch vehicles to the on-demand ride requests received in real-time or suggest relocations of empty vehicles, while ensuring the accepted scheduled requests in Stage 1 are satisfied. For expository purposes, we assume scheduled requests are received before all on-demand requests. However, our analysis and solution approach apply to a more general setting. While most work on trip-vehicle dispatch often ignores uncertainty in demand \cite{lee2004taxi,bertsimas2018online,alonso2017demand}, recent work starts to emphasize the uncertainties and the value of data \cite{alonso2017predictive,lowalekar2018online,moreira2013predicting,tong2017simpler,zhang2017taxi}. We also take a data-aware view and assume the platform knows the spatio-temporal distribution of the on-demand requests, which in practice can be estimated from historical data.

We propose new algorithms for both stages to handle both types of requests. In the design of these algorithms, we introduce a novel notion of Constrained Spatio-Temporal value function (CST-function). The CST-function is defined by construction with a polynomial-time algorithm we provide.
We show that CST-function represents the expected value a vehicle could gain under the optimal dispatch policy with the constraint that it needs to arrive at a specific location at a given time in the future. 
Built upon CST-function, we design a randomized best-fit algorithm for Stage 1 to decide whether to accept requests scheduled in advance in an online fashion. We also present theoretical bounds on the competitive ratio of algorithms for Stage 1 when there are no on-demand requests to be considered in Stage 2. In addition, we build an online planning algorithm for Stage 2 to dispatch vehicles to on-demand ride requests in real time given the accepted scheduled requests as constraints. This online algorithm runs in polynomial time with guaranteed optimality for the single-vehicle case. When multiple vehicles exist, the algorithm sequentially updates CST-function to dispatch the vehicles one by one. We demonstrate the effectiveness of the algorithms through extensive experiments on a real-world dataset of an online ride-hailing platform.

\section{RELATED WORK}

Trip-vehicle dispatch has been studied extensively but existing work only considers real-time on-demand requests or scheduled requests. 
With only scheduled requests, the problem is known as the Dial-a-Ride Problem (DARP)~\cite{cordeau2007dial,nedregaard2015integrated} and several variants of it have been studied ~\cite{cordeau2006branch,kim2011model,santos2015taxi,faye2016static,desaulniers2016exact,baldacci2012recent,chen2006dynamic}.
With only on-demand requests, dispatch algorithms use different approaches, such as greedy match \cite{lee2004taxi,bertsimas2018online}, 
collaborative dispatch~\cite{seow2010collaborative,zhang2016control,ma2013t}, planning and learning framework~\cite{xu2018large}, and receding horizon control approach~\cite{miao2016taxi}.
Our work is the first to consider both types of requests which lead to a two-stage decision-making process.

While our Stage 2 problem share similarities with the dispatch problem with on-demand requests only, the accepted scheduled requests in Stage 1 brought in hard constraints that cannot be handled easily. Simple extensions of existing algorithms \cite{xu2018large,lowalekar2018online} to our setting does not lead to a good performance as shown in our experiments. 

Our Stage 1 problem is closely related to the problem of online packing/covering~\cite{buchbinder2005online} , online Steiner tree~\cite{imase1991dynamic,awerbuch2004line}, online bipartite graph matching~\cite{karp1990optimal}, and the online flow control packing~\cite{garg2002line,buchbinder2009online}. Despite the similarities, 
none of those results or techniques could be directly applied to our setting due to the spatio-temporal constraints in our problem.

Other related work include efforts on last-mile transportation~\cite{cheng2014mechanisms,agussurja2018state,van2017delivery}, coordinating dispatching and pricing~\cite{chen2017optimal,Ma:2019:SPR:3328526.3329556,bai2018coordinating,fiat2018flow}, with a different focus of study.

\section{MODEL}
We consider a discrete-time, discrete-location model with single-capacity vehicles and impatient riders. We discuss relaxation of the assumptions in Section 6. 

Let $[T]=\{1,\ldots,T\}$ be the set of time steps, representing the discretized time horizon.
Let $[N]=\{1,\ldots,N\}$ be the set of different locations or regions. 
We denote by $\dist(u, v), u,v\in [N]$ the shortest time to travel from $u$ to $v$.
$\cD$ denotes the set of vehicles.
Each vehicle $c\in \cD$ is associated with time-location pairs $(\tilde{t}_c,\tilde{o}_c)$ and $(\tilde{t}'_c,\tilde{o}'_c)$, where $\tilde{t}_c\in [T]$ represents the earliest time that $c$ can be dispatched and $\tilde{o}_c\in [N]$ represents its initial location. Similarly, $(\tilde{t}'_c,\tilde{o}'_c)$ is defined to be the time-location pair representing when and where $c$ ends its service.

We employ a two-stage model for processing the scheduled requests and on-demand requests. 
In Stage 1, the system receives a sequence of scheduled requests. A scheduled request $r$ is described by a tuple $(o_r,d_r,t_r,v_r)$, representing a requested ride from the origin $o_r$ to the destination $d_r$ that needs to start at time $t_r$, and $v_r$ represents the \emph{value} the platform will receive for serving this request. 
We assume $v_r$ is given to the system when the request is made, e.g., provided by the rider or an external pricing scheme. 
When a scheduled request arrives, the platform should either accept and assign it to a specific driver or reject it. The decision needs to be made immediately before the next request arrives. We assume that the scheduled requests for a specific day will all arrive before the day starts.
We will discuss the relaxation of this assumption in Section 6. Let $\tilde{R}_c$ denote the set of accepted scheduled requests assigned to $c\in\cD$ by the end of Stage 1 and define $\tilde{R}=\cup\tilde{R}_c$.
W.l.o.g., we assume the vehicle always ends its service by serving a scheduled request in $\tilde{R}_c$. If not, a virtual scheduled request with request time and origin corresponding to $(\tilde{t}'_c,\tilde{o}'_c)$ could be added to $\tilde{R}_c$.

In Stage 2, the system starts to take real-time on-demand requests described by a tuple $(o_r,d_r,t_r,v_r)$ same as scheduled requests. 
We define the type of an on-demand request $r$ to be $(o_r,d_r,t_r)$ and let $\cW$ be the set of all possible types. 
We assume on-demand requests with type $w=(o,d,t)$ have the same value $V_w$ (or $V_{o,d,t}$) and $V_w$ $\forall w\in \cW$ is known to the system
in advance. This is a reasonable assumption if, for example, the variation in value for a trip of type $w$ is small in history and can be estimated from historical data.
Note that these requests are received in real-time, i.e., request $r$ will appear at time $t_r$. 
Upon receiving a set of on-demand requests $R_t$ at time $t$, the platform needs to decide immediately
for each request 
(1) either to dispatch it to an available vehicle $c$ currently at location $o_r$,
 in which case vehicle $c$ will start to serve the request and become available again at location $d_r$ at time $t_r + \dist(o_r, d_r)$, (2) or to reject this request. During the processing of these on-demand requests, the platform also needs to ensure that \emph{all} scheduled requests that it previously accepted 
must be served at their respective scheduled times and by their respective drivers.
In Stage 2, we also allow for relocating a vehicle to location $d$ when it is dispatched for no request. In this case, the vehicle will operate in the way as if it were taking a virtual request with destination $d$ and value 0. 

The goal of the system is to maximize the total value of all accepted requests, including both scheduled and on-demand requests. We do not assume any prior knowledge of the distribution of the scheduled requests due to the irregularity of scheduled requests, but we assume the distribution of real-time on-demand requests is known or can be estimated from historical data.
Let independent random variable $X_w$ denote the number of requests of type $w\in\cW$ that the system receives in Stage 2. The on-demand request distribution is described as $\Pr[X_w\geq i]$, for all $w\in \cW$ and $i\in \mathbb{N}$. 


\section{SOLUTION APPROACHES}
The system needs to deploy two algorithms for trip-vehicle dispatch, one for each stage.
Critically, the scheduled requests accepted in Stage 1 will serve as constraints in Stage 2. Consider the simplest setting where there is only one vehicle. 
If the system has accepted a scheduled request $r$, then in Stage 2, when the system receives an on-demand request $r'$ prior to the starting time of $r$, whether or not the system should dispatch the vehicle to serve $r'$ depends on (i) whether the vehicle can still serve $r$ after finishing the trip of $r'$, (ii) how much it can gain if it serves $r'$, (iii) how much it can gain if it does not serve $r'$.
In hindsight, whether or not the system should accept the scheduled request $r$ depends on the expected gain in Stage 2 with or without having $r$ as a constraint.

Based on this intuition, in this section, we will first introduce a novel notion of \emph{CST-function} which represents the expected gain of the only vehicle in the system in Stage 2 when it is committed to serve a scheduled request. We will then present the algorithm for Stage 2, followed by the algorithm for Stage 1, both built upon the CST-function.


\subsection{CST-FUNCTION}
The system's decision making problem in Stage 2 can be modeled as a Markov Decision Process (MDP) when there is only one vehicle in the system. Let $c$ be the only vehicle in the system. the system is faced with an MDP defined by $(\mathcal{S},\mathcal{A},\mathcal{P},\mathcal{V})$ where $\mathcal{S}$ is the set of states, $\mathcal{A}$ is the set of actions, $\mathcal{P}$ is the state transition probability matrix and $\mathcal{V}$ is the reward function. The set of accepted scheduled requests $\tilde{R}_c$ should be served reliably, and we encode this constraint in the definition of $\mathcal{S}$ and $\mathcal{A}$.

A state $s\in\mathcal{S}$ is defined by $(t,l,R_{t,l})$ where $t\leq \tilde{t}_c'$ is the time, $l$ is the location of the vehicle that is waiting to be dispatched, and 
$R_{t,l}$ is the set of currently received on-demand requests that can potentially be served by the vehicle while ensuring a reliable service for the scheduled requests. 
Given a request $r$, define $D(t,l|r):=\{d:d\in[N], \dist(l,d)+\dist(d,o_r)\leq t_r-t\}$ as the set of locations the vehicle could leave for from $l$ so that he will be able to reach $(o_r,t_r)$ after arriving at the location.
Then given $R_t$ and $\tilde{R}_c$, we have $R_{t,l}=\{r':r'\in R_t, o_{r'}=l, d_{r'}\in D(t,l|\tilde{r}_c^t), t_{r'}=t\}$ where $\tilde{r}_c^t\in R_c$ is the earliest scheduled request for $c$ with a pick-up time at or after time $t$.

Let $\mathcal{A}(s)$ be the set of available actions at state $s=(t,l,R_{t,l})$. If the pickup time of the request $\tilde{r}_c^t$ is $t$, then the only available action at $s$ is to assign the vehicle to $\tilde{r}_c^t$. Otherwise,
$\mathcal{A}(s)$ consists of two types of actions: assigning $c$ to a request $r'\in R_{t,l}$, or relocating $c$ to a location $l'\in D(t,l|\tilde{r}_c^t)$.
The state transition probability $\mathcal{P}^a_{ss'}=\Pr[S_{\tau+1}=s'|S_{\tau}=s,A_{\tau}=a]$ is non-zero only when the vehicle becomes available again at $(t',l')$ of $s'$ after taking action $a$ at $s$ and
$\mathcal{P}^a_{ss'}=\prod_{w\in\cW}\Pr[X_w= X_{s',w}]$, where $X_{s',w}$ is the number of type-$w$ requests in $R_{t',l'}$ at state $s'$. The immediate reward $\mathcal{V}_{s}^a$ is $v_r$ if $a$ corresponds to dispatching $c$ to a scheduled or on-demand request $r$, and $\mathcal{V}_{s}^a=0$ otherwise. 
The state value function and state action value function under a policy $\pi$ are denoted by $v_{\pi}(s)$ and $q_{\pi}(s,a)$ respectively. The total reward of the MDP is regarded as the sum of $\mathcal{V}_s^a$, without applying a discount factor.

Given a vehicle at $(t,l)$ with with $r$ ($t_r\geq t$) being the next request it is required to serve, we define the CST-function $\f(t,l|r)$ by an algorithm to compute it as shown in Algorithm \ref{Algo1}. 
Instead of going through the algorithm, we will first show an important property of the CST-function.

\begin{lemma}\label{lemma1}
	If $\tilde{R}_c$ contains only one (real or virtual) request $r$,
	then $\f(\cdot)$ computed by Algorithm 1 satisfies
	\[\f(t,l|r)=\mathbb{E}_{\mathcal{R}_{t,l}}\left[v_{\pi^*}(t,l,\mathcal{R}_{t,l})\right]\] where $\pi^*$ is the optimal policy and it follows \[q_{\pi^*}(s,a)=\mathcal{V}_s^a+\f(t_a,l_a|r),\] where $(t_a,l_a)$ is the time-location pair that action $a\in \mathcal{A}(s)$ will lead to when the vehicle becomes available again.
\end{lemma}

The detailed proof is deferred to Appendix \ref{proofLem1}.
By lemma \ref{lemma1}, $\f(t,l|r)$ is the weighted average state value of states with time $t$ and location $l$ and varying $R_{t,l}$. Therefore, it represents the expected value a vehicle at $(t,l)$ could gain before it reaches $r$ under the optimal policy. 
Thus the recursive algorithm shown in Algorithm \ref{Algo1} can be interpreted as follows. 
It first calculates $\f(\cdot)$ for relevant future time-location pairs (line 1-4), then determines an ordered list of destination locations $a_1,\ldots,a_j,d^*$ worth considering (line 5-7) and $d^*$ is the location with highest CST value if driving idly (line 6), 
which encodes the system's preferences over requests. If there is a request to $a_i$ but no request to $a_k$ $\forall k<i$, the request to $a_i$ will be served, leading to an immediate reward $V_{l,a_i,t}$ and an expected future gain of $\f(t+\delta(l,a_i),a_i|r)$. 
The algorithm computes $\f(t,l|r)$ based on the probability that each of these events happens and the corresponding reward (line 9-12). The system will never consider certain requests since guiding the vehicle to drive idly towards $d^*$ is more promising (line 13).

Now we can claim that $\f(t,l|r)$ represents the expected total value of trips a vehicle can serve between time $t$ and the start time of $r$ given (i) it is located at $l$ and is available to serve an on-demand request at time $t$; (iii) it is committed to serve $r$ in the near future; (iv) it is the only vehicle in the system.
In fact, the CST-function is in concept similar to the state value function and Q-function (state-action value function) in sequential decision making \cite{howard1960dynamic}, but it is specially designed for our problem with two key features. First, it considers the constraint due to Stage 1 in our problem. Second, CST-function is more compact than state value function and Q-function in our problem. Notice that the MDP of our problem has an exponential number of states as $R_{t,l}$ can take any subsets of the potential rides starting from time $t$ at location $l$. Therefore, computing either state value function or Q-value function would be inefficient in both memory and computation time. In contrast, CST-values are only relevant to the vehicle's location and time and we only need polynomial-sized space to store the values and it can be computed in polynomial time.


\subsection{SOLVING STAGE 2}

In this section, we present our solution approach for Stage 2.
When there is only one vehicle in the system, we design an algorithm DPDA (Dynamic Programming based Dispatch Algorithm) with the aid of the CST-function, and prove that it induces the optimal policy for the MDP. The algorithm can be extended to multiple-driver setting  to find the optimal policy,
but it requires exponential memory and runtime since it is necessary to include the time-location pairs of all vehicles in the states used in dynamic programming. Therefore, we provide an alternative algorithm DPDA-SU (DPDA with Sequential Update) that extends DPDA by sequentially dispatching available vehicles and updating a virtual demand distribution. 


\paragraph{Single-Vehicle Case}

To solve this MDP, we introduce the DPDA algorithm, which implicitly induces a policy for the MDP.   As shown in Algorithm \ref{Algo2}, DPDA suggests a way to make the online decision for the vehicle given its current state $s=(t,l,\cR_{t,l})$ and $\tilde{r}_c^t$, with the aid of the CST-function $\f(t,l|r)$. 
It first calculates the CST-function  $\f(t_a,l_a|\tilde{r}_c^t)$ for all $a\in \mathcal{A}(s)$ (line 2-4) and chooses the action with the highest expected value the vehicle could gain before reaching $\tilde{r}_c^t$ (line 5). 
Next, we show that  Algorithm \ref{Algo2} induces an optimal policy for the MDP.

\begin{algorithm}[t]
	\caption{Calculate $\f(t,l|r)$}\label{Algo1}
	\begin{algorithmic}[1]
		\If {$l=o_r$ and $t=t_r$}
		\State \Return $0$
		\EndIf
		\For {$d\in D(t,l|r)$}
		\State Calculate $\f(t+\dist(l,d),d|r)$
		\EndFor
		\State Denote $\{a_i\}$ the sequence of $d\in D(t,l|r)$ in decreasing order of $V_{l,d,t}+\f(t+\dist(l,d),d|r)$
		\State $d^*\gets\arg\max_{d\in D(t,l)} \f(t+\dist(l,d),d|r)$
		\State  $j$ $\gets$ the largest index of $\{a_i\}$ such that $V_{l,a_j,t}+\f(t+\dist(l,a_j),a_j|r)>\f(t+\dist(l,d^*),d^*|r)$
		\State $p\gets 1$
		\State $F\gets 0$
		\For {$i=1$ to $j$}
		\State $F\gets F+p\cdot \Pr[X_{l,a_i,t}\geq 1](V_{l,a_i,t}+\f(t+\dist(l,a_i),a_i|r))$
		\State $p\gets p\cdot (1-\Pr[X_{l,a_i,t}\geq 1])$
		\EndFor
		\State $F\gets F+p\cdot \f(t+\dist(l,d^*,d^*)|r)$
		\State $\f(t,l|r) \gets F$
	\end{algorithmic}
\end{algorithm}

\begin{algorithm}[t]
	\caption{DPDA$(s=(t,l,\mathcal{R}_{t,l})|\tilde{r}_c^t)$ }\label{Algo2}
	\begin{algorithmic}[1]
		\State Determine the action set $\mathcal{A}(s)$
		\For {$a\in \mathcal{A}(s)$}
		\State $(t_a,l_a,)\gets$ the time-location pair action $a$ leads to
		\State Calculate $\f(t_a,l_a|\tilde{r}_c^t)$
		\EndFor
		\State $a^*=\arg\max_{a\in \mathcal{A}(s)}\mathcal{V}_{s}^a+\f(t_a,l_a|\tilde{r}_c^t)$
		\State \Return $a^*$
	\end{algorithmic}
\end{algorithm}

\begin{theorem}\label{thm1}
	Algorithm \ref{Algo2} induces an optimal policy.
\end{theorem}

\begin{proof-sketch}
	We claim that the optimal policy $\pi^*$ for state $s=(t_c,l_c,\cR_{t_c,l_c})$ should only depend on $\tilde{r}_c^t$.
	Thus the overall MDP can be decomposed into several local MDPs with respect to each of the scheduled requests in $\tilde{R}_c$. Hence by Lemma \ref{lemma1}, solving the overall MDP is equivalent to solving the local MDP corresponding to $\tilde{r}_c^t$ (line 5 in Algorithm \ref{Algo2}), which concludes the proof.
\end{proof-sketch}


\paragraph{Multi-Vehicle Case}

To compute the optimal solution for the multi-vehicle case at time step $t$, a dynamic programming that is similar to Algorithm \ref{Algo1} could still be applied. However, it suffers from the curse of dimensionality, which will lead to an exponential algorithm regarding time complexity and space complexity.
To circumvent the difficulty, we provide a heuristic sequential algorithm  as follows, as well the intuition behind. We start from the case with two vehicles $c_1$ and $c_2$. For the first vehicle $c_1$, we treat it as if it were the only vehicle in the system and we  decide an action $a^*$ for $c_1$ by running the DPDA. Let $p_w$ be the probability a request of type $w\in\cW$ being served by this vehicle given $X_w\geq 1$, and notice that $p_w$ could be obtained during the computation of the CST value. Afterward, we could obtain a new marginal distribution of on-demand requests. That is, given $a^*$, for any $i\in\mathbb{N}$ we have 
\begin{equation}\label{eq1}
\begin{split}
\Pr[X'_w\geq i|a_{c_1}=a^*]  = \,& (1-p_w)\cdot\Pr[X_w\geq i] \\
& + \,p_w\cdot\Pr[X_w\geq i+1] 
\end{split}
\end{equation}
where random variable $X'_w$ denotes the number of remaining requests of type $w$.
Then for the second vehicle, we run the DPDA again as if it were the only vehicle and use the updated marginal distribution as the new distribution of on-demand requests. 

For the case with more than 2 vehicles, we dispatch orders sequentially for each vehicle by simply repeating the procedure described above. That is, we sequentially run DPDA for each vehicle and update a virtual demand distribution. 
Note that after the second vehicle, the marginal distribution we maintained is not accurate anymore since we ignored their potential correlation. Nevertheless, it could serve as a reasonable estimation of the actual probability for our algorithm. In the description below, we use function $h(\cdot)$ to denote this estimated marginal distribution.

Following the intuition described above, we formally introduce the DPDA-SU in Algorithm \ref{Algo3}. 
In the multi-vehicle case, as shown in Algorithm 3, for each vehicle we sequentially run DPDA (line 6) and update a virtual demand distribution represented by $h(X_w\geq i)$ (line 7) and recompute the CST-function (line 6) assuming $\Pr[X_w>i]=h(X_w\geq i)$, the updated virtual demand distribution, after each call of DPDA. Indeed, $h$ serves as an approximation of the updated marginal probability $\Pr[X_w\geq i]$ after a vehicle is assigned to a ride request. Note that when a vehicle is assigned to a request, it not only changes the virtual distribution of trips starting from the current time step, but also in the future time steps because the assigned vehicle can serve future demands after it completes the current ride. So the key is in the update of $h$, which is done following equation (\ref{eq1}). Intuitively, we first get $p_w$, the probability that a request of type $w$ will be served by the vehicle $c$ which is just assigned (in the last iteration) to a ride request, assuming it is the only vehicle in the system, and then update the distribution. $p_w$ is in fact a byproduct of the computation of CST-function in the last iteration. 
We defer the pseudocode of the distribution update to Appendix \ref{AlgoUpdateDist}. 

In addition, in line 2 of Algorithm \ref{Algo2}, we do not fix the choice of vehicle sequences. In experiments, we investigate how the choice of vehicle sequences impact the outcome, specifically the variance of values gained by each vehicle, since the variance relates to the fairness of a dispatching algorithm which is a practical concern in many ride-hailing platforms with self-interested drivers.

\begin{algorithm}[t]
	\caption{DPDA-SU}\label{Algo3}
	\begin{algorithmic}[1]
		\State Get the probabilities $\Pr[X_w\geq i]$ and value $V_{w}$ for all $w\in\mathcal{W}$ from historical data
		\For{$w\in\cW$, $i\in \mathbb{N}$}
		\State $h(X_w\geq i)\gets \Pr[X_w\geq i]$
		\EndFor
		\For{$c\in \mathcal{D}$}
		\State $r_c\gets$ next scheduled request for $c$
		\State  $a_c\gets $DPDA$((t_c,l_c,\mathcal{R}_{t_c,l_c})|\tilde{r_c})$ with $h(\cdot)$ as $\Pr[\cdot]$
		\State $h(\cdot)\gets$ \UpdateDist$(h(\cdot),a_c,r_c)$
		\EndFor
	\end{algorithmic}
\end{algorithm}

\subsection{SOLVING STAGE 1}
In this section, we design and analyze request selection algorithms for Stage 1. 
In this stage, the platform receives a sequence of scheduled requests and needs to decide their assignments in an online fashion. These requests are all received before any of the on-demand requests. 






We aim to design efficient online selection algorithms for Stage 1. 
To evaluate the performance of such online algorithms, we employ the notion of \emph{competitive ratio}, which is a commonly used notion in online algorithm analysis. Given an input instance $\mathcal{I}$, we denote $\OPT(\mathcal{I})$ and $\ALG(\mathcal{I})$ as the optimal offline solution and the solution of an online algorithm on $\mathcal{I}$. We say the online algorithm is $\gamma$-competitive, if $\frac{\mathbb{E}[\OPT(\mathcal{I})]}{\mathbb{E}[\ALG(\mathcal{I})]}\leq \gamma$ holds for every problem instance $\mathcal{I}$.

As a start, we focus on Stage 1 problem alone without any interference from Stage 2. That is, 
we first assume that there is no on-demand request in Stage 2, and the goal of the selection algorithm is to select a set of feasible scheduled requests with maximum total value. We further assume that the value $v_r$ of any scheduled request $r$ is proportional to the trip distance. Thus, w.l.o.g, we simply set $v_r$ equal to $\dist(o_r,d_r)$. 
In this setting, we show a tight competitive ratio on any deterministic online algorithms. This ratio depends on a parameter $\mu$, which is defined to be the ratio between the largest and smallest value of all possible requests.


\begin{theorem}\label{thmworstcase}
	If $v_r=\dist(o_r,d_r)$ and there is no on-demand request in Stage 2, any deterministic online algorithm for Stage 1 has a competitive ratio at least $4\mu-1$.
\end{theorem}


Next, we show that a simple first-fit algorithm that always dispatches requests to the first available vehicle if there exists one is $4\mu-1$ competitive, proving that the bound is tight.

\begin{algo*}[\FirstFit]
Fix an arbitrary order of the vehicles. For each incoming scheduled request, always assign it to the first vehicle in order that could serve this request without any conflicts. If no such vehicle exists, reject this request.
\end{algo*}

\begin{theorem}\label{thmgreedy}
	If $v_r=\dist(o_r,d_r)$ and there is no on-demand request in Stage 2, algorithm \FirstFit for Stage 1 is $(4\mu-1)$-competitive.
\end{theorem}

\newcommand{\rank}{{\mathsf{rank}}}

The proofs of Theorem  \ref{thmworstcase} and \ref{thmgreedy} are deferred to Appendix.
Next, we take into consideration the Stage 2 on-demand requests, which are assumed to follow the distribution $Pr[X_w \geq i]\,\forall w\in\mathcal{W}, i\in\mathbb{N}$.  

First, upon the arrival of a scheduled request $r$, for each vehicle $c$ that can serve $r$, we estimate the expected value increment from this assignment with the help of the CST-function $\f(t,l|r)$.
More specifically, let $r_0$ and $r_1$ be the accepted scheduled requests vehicle $c$ serves before and after $r$ (if $r_0$ or $r_1$ does not exist, we set a virtual request that corresponds to the start or end time-location pair of vehicle $c$). Then we set
\begin{eqnarray*}
	E_0=\f(t_{r_0}+\dist(o_{r_0},d_{r_0}),d_{r_0}\mid r_1)
\end{eqnarray*}
to be the estimated value of vehicle $c$ without taking request $r$, and 
\begin{eqnarray*}
	E_1=&\f(t_{r_0}+\dist(o_{r_0},d_{r_0}),d_{r_0}\mid r)\\
	&\,+v_r+\f(t_{r}+\dist(o_{r},d_{r}),d_{r}\mid r1)
\end{eqnarray*}
to be the estimated value of vehicle $c$ after taking request $r$. We then define the estimated value increment of request $r$ for vehicle $c$ as $\Delta_c(r) = E_1-E_0$.
Such estimation suggests a greedy algorithm as follow.

\begin{algo*}[\BestFit]
For each coming scheduled request $r$,
assign it to the vehicle $c$ with which serving $r$ could yield the highest value increment $\Delta_c(r)$; if no vehicle could serve $r$, reject this request.
\end{algo*}

Note that in this algorithm, we do not reject any requests as long as there are vehicles that can serve it and $\Delta_c(r)$ could be negative. As we will see in the experiments in Section 6, variants of \BestFit that treat the case where $\Delta_c(r)\leq 0$ differently result in lower performance than the original algorithm.

Finally, in the last algorithm, we add an additional random priority component to the value increment. 
Inspired by the online bipartite graph matching algorithm proposed by~\cite{karp1990optimal}, we assign each vehicle a weight $\rank(k)=e^{\alpha k}$ that denotes its priority, where $k$ is a random variable drawn from the uniform distribution $U[0,1]$ independently for each vehicle and $\alpha$ is a constant. The new estimated value increment of vehicle $c$ serving request $r$ then becomes $\Delta'_c(r) = E_1 - E_0 + \beta * e^{\alpha k}$, where $\beta$ is another scaling parameter.

Our final algorithm, \RandomBestFit, choose the vehicle based on this newly randomized value increment.



\begin{algo*}[\RandomBestFit]
For each coming scheduled request $r$,
assign it to the vehicle $c$ with which serving $r$ could yield the highest randomized value increment $\Delta'_c(r)$; if no vehicle could serve $r$, reject this request.
\end{algo*}

\section{EXPERIMENTS}
In this section, we demonstrate the effectiveness of the proposed algorithms. First, we introduce the dataset and describe how we process and extract information from it. Then we introduce baseline algorithms for both stages and present the experimental results.

\subsection{DATA DESCRIPTION AND PROCESSING}
We perform our empirical analysis based on a dataset provided by Didichuxing.
The dataset consists of $2\times 10^5$ valid requests. Each request $r$ consists of the start time $t_r$, the duration of the trip, the origin $o_r$, the destination $d_r$, and its assigned vehicle ID. The value $v_r$ is not given from the dataset, and we set it to be proportional to the duration of the trip.

The  locations in the dataset are represented by latitudes and longitudes. We transform them into discretized regions by running a $k$-means clustering algorithm on all the valid coordinates.
We obtain 21 centers after 61 rounds of iteration (details in Appendix). 
Then the discretized label of each location in the dataset is represented by the label of its nearest center and $\dist(o,d)$ are calculated based on the coordinates of centers of regions $o,d$.
The time horizon is discretized into 1 minute per time step.
Finally, the distribution of on-demand requests from the data can be derived given the discretized time horizon and regions.
For each vehicle $c$, $(\tilde{t_c},\tilde{o_c})$ are used as  the earliest occurrence time and location of $c$ given in the dataset. We set $(\tilde{t'_c},\tilde{o'_c})=(\tilde{t_c},\tilde{o_c})$.




\subsection{EXPERIMENT SETUP}
All experiments are done on an i7-6900K@3.20GHz CPU with 128GB memory.
We introduce the default global setup for all the experiments. 
The duration of a time step is set to 1 minute and we have $24\times 60=1440$ time steps in each iteration. 
Next, we sample on-demand requests from the historical distribution derived from the dataset, with an average number of 1804 generated requests in each iteration. 
For scheduled requests in Stage 1, we set their frequency to be $1/20$ of that of the on-demand requests in Stage 2, and the types of 87 scheduled requests are drawn i.i.d. from $\cW$ following the on-demand requests distribution. The value of a request of type $w$ is set to be $V_{w}$.
Finally, a set of 50 vehicles are drawn uniformly from the dataset. All experiments below follow this setup unless specified otherwise.





\subsection{BASELINE ALGORITHM}
We compare our algorithms with several baseline algorithms for both Stage 1 and Stage 2.
For Stage 1, we employ the First-Fit algorithm as the baseline.
For Stage 2, we employ two matching based algorithms (Greedy-KM and Enhanced KM), a learning and planning based algorithm (LPA), and a sampling-based mixed integer linear programming (S-MILP) algorithm.
Greedy-KM dispatches requests myopically considering only their values.
Enhanced KM is an extension of Greedy-KM with the CST value. 
The LPA is adapted from \cite{xu2018large} to handle the hard constraints brought in by the scheduled requests and we implement it with slight changes of the setting in \cite{xu2018large}. 
In \cite{lowalekar2018online}, assignments between vehicles and riders  at time step $t$ are made by solving a MILP that takes into account several samples of requests at the time step $t+1$. The S-MILP is an extension of \cite{lowalekar2018online} by adding scheduled requests as constraints in the MILP. In our experiments, the number of samples is set to 10.
The details of those baseline algorithms are provided in Appendix.

\subsection{RESULTS}
\begin{figure*}[htbp]
	\centering
	\begin{subfigure}[t]{0.32\textwidth}
		\centering
		\includegraphics[width=5cm]{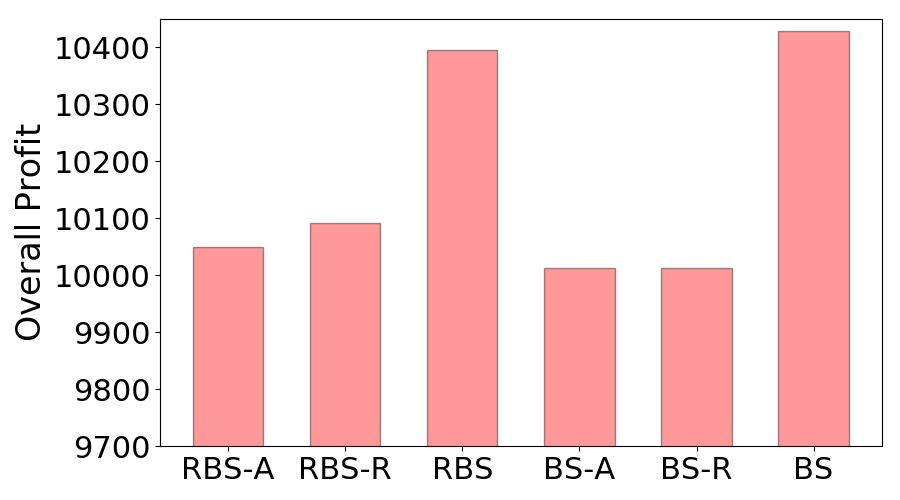}
		\caption{Performances of Stage 1 Algorithms.}
		\label{figheur}
	\end{subfigure}
	~	
	\begin{subfigure}[t]{0.32\textwidth}
		\centering
		\includegraphics[width=5cm]{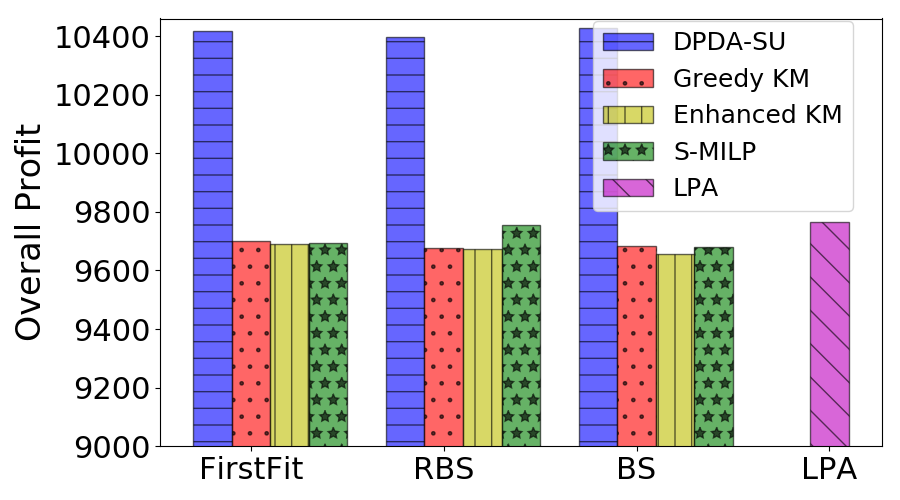}
		\caption{Performance of Pairwise Combinations and the LPA.}
		\label{figall}
	\end{subfigure}
	~
	\begin{subfigure}[t]{0.32\textwidth}
		\centering
		\includegraphics[width=5cm]{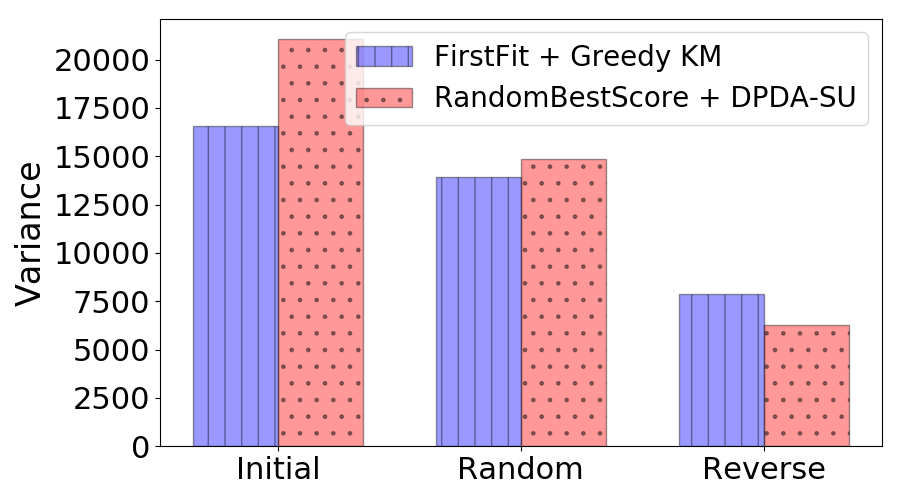}
		\caption{Variances when Using Different Sequence Choices.}
		\label{fig_var}
	\end{subfigure}

	\begin{subfigure}[]{1.0\textwidth}
		\centering
 		\includegraphics[width=16cm]{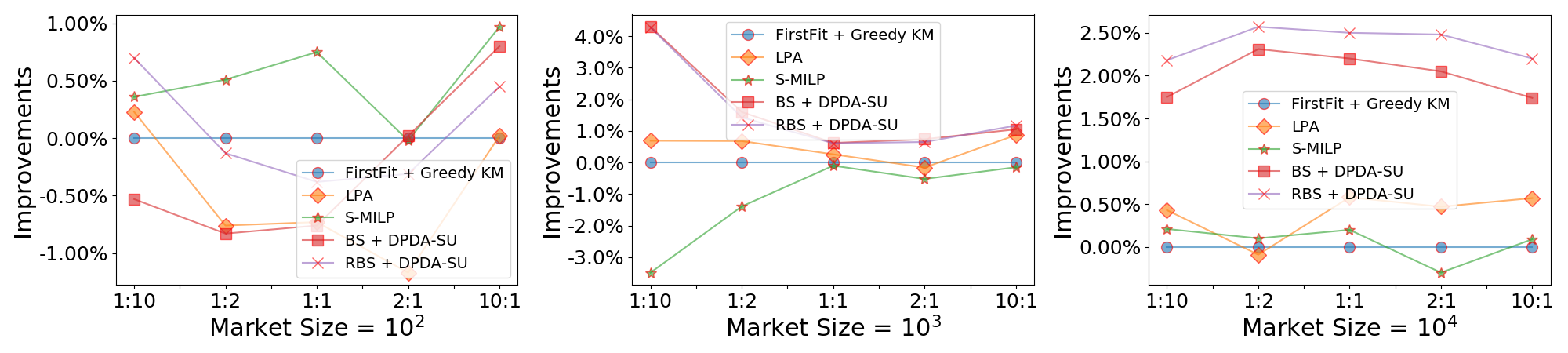}
		\caption{Improvement w.r.t. the Baseline in Different Market Sizes.}
		\label{fig_size}
	\end{subfigure}	
	\caption{Experimental Results}
	\label{Exp}
\end{figure*}

First, we combine DPDA-SU for Stage 2 with \BestFit (BS), \RandomBestFit (RBS) and their variants for Stage 1, to evaluate the performance of all combinations of our proposed algorithms. We consider the following two variants of \BestFit: (1) accept a request only if the highest incremented value $\Delta_c(r)$ is positive; (2) if the highest incremented value $\Delta_c(r)$ is negative, accept request $r$ and assign it to $c$ with probability $e^{\Delta_c(r)}$. We denote these two variants as \BestFitR and \BestFitA, respectively. We also define the two variants of \RandomBestFit, \RandomBestFitR and \RandomBestFitA, in a similar way  to the two variants of \BestFit. 
For the multi-vehicle case, the results are shown in Figure \ref{figheur}. 
One can see that each of \RandomBestFit and \BestFit outperforms its variants significantly. Thus, in the rest of the experiments, we employ \RandomBestFit and \BestFit as  our Stage 1 algorithms. We also provide additional results for the single-vehicle case in Appendix~\ref{append:SV}.
Next, we conduct experiments on pairwise combinations of Stage 1 and Stage 2 algorithms, as well as the LPA. The results are shown in Figure \ref{figall}.
We can conclude that when one of \FirstFit, \BestFit, \RandomBestFit is fixed for Stage 1, DPDA-SU always outperforms Greedy-KM, Enhanced-KM, and S-MILP.
Though the LPA outperforms the other combinations without DPDA-SU, those with DPDA-SU are significantly better than the LPA.
We further test our algorithms by varying the market parameters. We test with markets of different numbers of requests of $10^2$, $10^3$, $10^4$, and different ratios $\kappa$ between the numbers of scheduled and on-demand requests. 
We deploy \FirstFit with Greedy-KM as one of the baselines. The reason we do not choose \FirstFit with Enhanced-KM is that Greedy-KM outperforms Enhanced-KM in all combinations as shown in Figure \ref{figall}. We  choose the LPA and S-MILP as the other two baselines.

Figure \ref{fig_size} shows the increase of profit of our algorithms compared to the baselines.
In the small market of $100$ requests, the baselines perform better than our algorithms in some cases. 
However, the significance test shows the $p$-values are significantly larger than $0.1$ in all cases in this market, which means no statistical conclusion can be drawn from these experiments.
For larger markets of $10^3$ and $10^4$ requests, our algorithms are on average better than the baselines for every $\kappa$. We also conduct the significance test in each market and the $p$-values in all cases are less than $10^{-6}$. Thus one can statistically conclude that our algorithm outperforms the baselines in large markets.

To verify the effectiveness of Stage 1 algorithms, 
we empirically compute the competitive ratios under the setting with only scheduled requests for each of \FirstFit, \BestFit, and \RandomBestFit.
We generate 50 instances, each with 87 scheduled requests.
For each algorithm $\ALG$, we compute $\frac{\mathbb{E}[\OPT(I)]}{\mathbb{E}[\ALG(I)]}$ for each instance $\mathcal{I}$ and the maximum is taken over all 50 instances as the empirical competitive ratio for $\ALG$.
For \RandomBestFit which is a randomized algorithm, we run the algorithm on each  instance 50 times and take the average output value as the estimate of $\mathbb{E}[\text{\RandomBestFit}(I)]$.
The offline optimal value for each instance $\OPT(\mathcal{I})$ can be calculated by a flow-based approach, as described in Appendix \ref{FlowBased}. The empirical ratios are summarized in Table \ref{tab:cr}, which shows that our algorithms have relatively low empirical competitive ratio compared to \FirstFit. This suggests that \BestFit and \RandomBestFit are  good candidate algorithms for markets with only scheduled requests.

\begin{table}[!h]
	\centering
		\begin{scriptsize}
	\begin{tabular}{c|c}
		\hline
		Algorithms in Stage 1 & Competitive Ratio \\ 
		\hline
		\BestFit & \textbf{1.38609} \\
		\hline
		\RandomBestFit & 1.39112 \\
		\hline
		\FirstFit &	1.4454 \\
	\end{tabular}
	\end{scriptsize}
	\caption{Stage 1 Competitive Ratios of Different Models.}
	\label{tab:cr}
\end{table}

In addition to the overall profit, we also test the variance of values gained by each vehicle with our algorithms. 
We consider the choice of vehicle sequences before running DPDA-SU that could lead to low variances without harming the total value. 
We test three variations.
In the first one, denoted as $\mathsf{Initial}$, we fix a vehicle order. In the second one,
denoted as $\mathsf{Reverse}$, we sort the vehicles in increasing order of the values they have already gained before running DPDA-SU. In the last variation, denoted as $\mathsf{Random}$, we shuffle the vehicles randomly. 
When using these three variations in the same algorithm, the differences in the total value are within 0.60\% from each other. 
The variance results are shown in Figure \ref{fig_var}.
One can see that when applying $\mathsf{Reverse}$, our algorithm \RandomBestFit with DPDA-SU leads to a lower variance than \FirstFit combined with Greedy-KM.
\begin{table}[t]
	\centering
	\begin{scriptsize}
	\begin{tabular}{c|c|c}
		\hline
	       &\textbf{DPDA-SU+\BestFit} & LPA \\
	    \hline
         Stage 1 & 2.2\% & 3.3\% \\
         \hline
         Stage 2 & 52.3\% & 56.0\% \\
	    \hline
        Overall & 50.0\% &  53.5\% \\
	    \hline
	\end{tabular}
		\end{scriptsize}

	\caption{Reject Rates of Different Stages.}
	\label{tab:rejection}
\end{table}

 We also investigate how the CST-value changes as the number of vehicles increases. We provide the details and result in  Appendix \ref{append:CSTVN}.

In real world, it could be bad service to reject the scheduled requests, so we evaluate the index of reject rate of DPDA-SU with \BestFit and the LPA. In Table~\ref{tab:rejection}, we show that in Stage 1 and overall,  
the reject rates of our algorithm is lower than the LPA.

Finally, we evaluate the scalability of Stage 2 algorithms in terms of their space complexities and running times.
The space complexities of different algorithms are summarized in Table~\ref{tab:rtime}. Here we denote $|\cD|$ as the number of vehicles, $M$ as the total requests, $m$ as the maximum number of requests at one time step, $N$ as the number of regions on the map, $T$ as the total time steps, and $\theta$ as the sampling times for only S-MILP algorithm ( $\theta=10$ in experiments).
Because most of the baseline algorithms are heuristics or mixed integer linear programs, it is hard to analyze their theoretical time complexities. Instead, we evaluate the running times of these algorithms in a fixed time window with different numbers of vehicles.
To test the most stressful situation, following the derived distribution, one time step with the most serious congestion is selected and amplified. 
On an average, 519 on-demand requests are generated for each time step. We assume no scheduled requests in Stage 1. 
For the markets with 1000, 3000, 5000 idle vehicles, we test the running time respectively, and the results are summarized in Table~\ref{tab:rtime}.
Though S-MILP has the shortest running time, when the number of vehicles increases, memory soon becomes a bottleneck for S-MILP. This is because the space required for S-MILP is quadratic in the number of vehicles. In our experiments, S-MILP runs out of memory when the number of vehicles reaches 5200 or higher.
On the other hand, the space required for DPDA-SU is linear in the number of vehicles. As a result, our algorithm can handle much larger markets than S-MILP. 
We then increase the number of regions to 200 and obtain the centers of each region using the same clustering algorithm.
Following the request distribution, we generate 1128 on-demand requests at each time step within the rush hour (9PM - 10PM). In this case, our algorithms can compute the results for each time step within 0.35 seconds using 115.2GB of memory.



\begin{table}[t]
	\centering
	\begin{scriptsize}

	\begin{tabular}{c|ccc|c}
		\hline
		Vehicles & 1000 & 3000 & 5000 & Space Complexity \\ 
		\hline
		S-MILP & 3.3 & 4.6 & 10.1 & $\mathcal{O}(|\cD|m\theta \cdot \max(|\cD|, m))$\\
 		\hline
		\textbf{DPDA-SU} & \textbf{10.4} & \textbf{31.6} & \textbf{56.2} & $\mathcal{O}(N^2T|\cD|)$\\
		\hline
		Greedy-KM & 1.7 & 32.0 & 133.7 & $\mathcal{O}(|\cD|m)$\\
		\hline
		LPA & 2.7 & 35.2 & 147.1 & $\mathcal{O}(\max(|\cD|m, MT))$ \\
		\hline
		Enhanced-KM & 9.5 & 56 & 185.3 &  $\mathcal{O}(N^2T|\cD| + |\cD|m))$ \\
		\hline
	\end{tabular}
	\end{scriptsize}
	\caption{Running Time (in seconds) with Different Number of Vehicles and Space Complexities of Different Algorithms.}
	\label{tab:rtime}
\end{table}
\section{CONCLUSION AND FUTURE WORK}
In this paper, we investigated the problem of trip-vehicle dispatch with the presence of scheduled and on-demand request. We proposed a novel two-stage model and novel algorithms for both stages. Through extensive experiments, we demonstrated the effectiveness of the algorithms for real-world applications.

The model can be applied to or further extended for problems with relaxed assumptions. First, our work can be applied to problems with patient requests, which can be treated as duplicated requests when there is only one driver. Second, our framework can be extended to the case where each scheduled request becomes available at least $\mu$ time before departure, where $\mu$ is the longest possible trip time. In this case, at each time step, we first deal with the newly-received scheduled requests before processing the on-demand requests and computing the CST-function. Third,  our algorithms can also deal with uncertainties in travel time, i.e., $\dist(u,v)$'s are not the same at different time steps. We could handle these uncertainties by replacing $\dist(u,v)$ with $\dist(u,v,t)$ in the algorithms, where $\dist(u,v,t)$ is the shortest time to travel from $u$ to $v$ that depends on $t$. For further investigation, our work can be integrated with work on last-mile routing to handle actual road networks.

\subsubsection*{Acknowledgement}
Fei Fang is funded in part by Carnegie Mellon University’s Mobility21, a National University Transportation Center for Mobility sponsored by the US Department of Transportation. The contents of this report reflect the views of the authors only.


\newpage

\bibliographystyle{mystyle}
\bibliography{ref}  
\newpage
\begin{appendices}
	
	\section{Proof of Lemma \ref{lemma1}}\label{proofLem1}
\begin{figure}[ht!]
		\begin{center}
			\includegraphics[width=4cm]{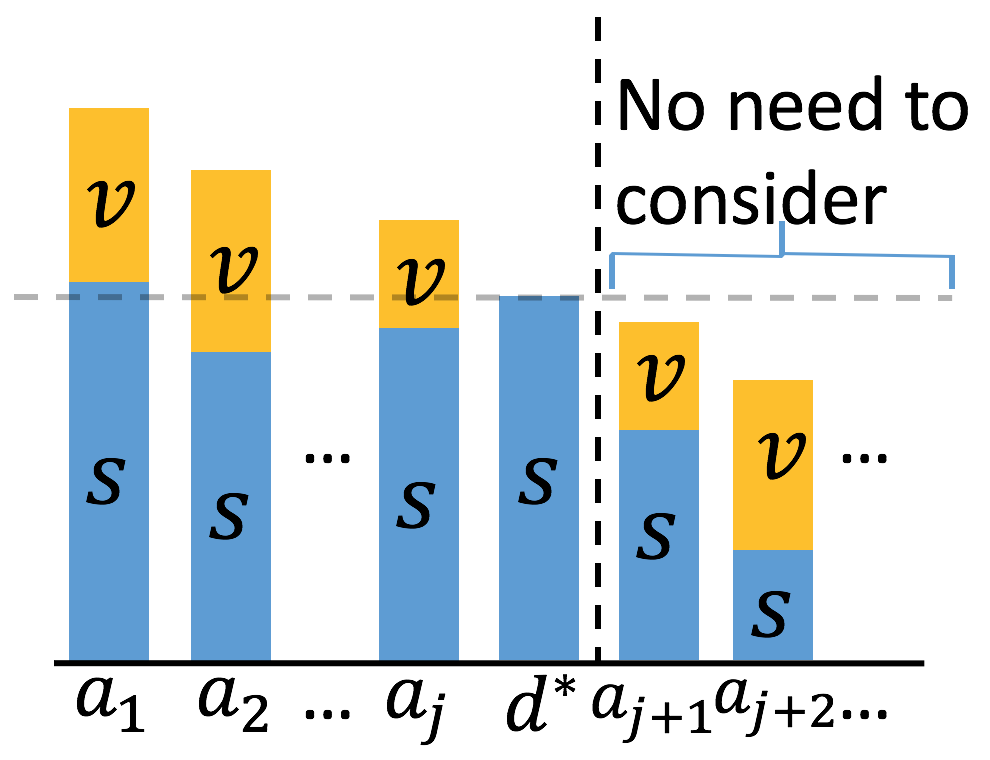}

			\caption{An Illustration of Algorithm \ref{Algo1}. In the bar for $a_i$, $\mathsf{s}$ represents $\f(t+\dist(l,a_i))$ and $\mathsf{v}$ represents $V_{l,a_i,t}$. }
			\label{figthm1}
		\end{center}
		
	\end{figure}

	\begin{proof}
		
				 By definition, $c$ ends its service upon completing $r$.
First we build up a transition graph $G'=(\mathcal{S},E')$ among the set of possible states in the corresponding MDP, where $E'=\{(s,s')|\exists a\in\mathcal{A} \text{ s.t. } \mathcal{P}_{ss'}^a>0\}.$ Clearly, $G'$ is a directed acyclic graph (DAG). 
		Thus from Bellman Expectation Equation \[v_{\pi}(s)=\sum_{a\in\mathcal{A}}\pi(a|s)\left(\mathcal{V}^a_s+\sum_{s'\in\mathcal{S}}\mathcal{P}_{ss'}^a{v}_{\pi}(s')\right),\] we can see that there exists an optimal deterministic policy, $\pi(a^*|s)=1$ where 
		\begin{equation}\label{eqa}
		\begin{split}
		a^*&=\arg\max_{a\in \mathcal{A}(s)}\mathcal{V}^{a}_s+\sum_{s'\in\mathcal{S}}\mathcal{P}_{ss'}^{a}{v}_{\pi}(s')\\
		&=\arg\max_{a\in \mathcal{A}(s)}\mathcal{V}^{a}_s+\mathbb{E}_{\mathcal{R}_{t_a,l_a}}\left[v_{\pi}(t_a,l_a,\mathcal{R}_{t_a,l_a})\right],
		\end{split}
		\end{equation}             
		and the state value together with the optimal policy can be determined following the topological ordering of states in $G'.$ 
		
		Next we show that $\f(t,l|r)=\mathbb{E}_{\mathcal{R}_{t,l}}\left[v_{\pi}(t,l,\mathcal{R}_{t,l})\right]$ by induction on $t$.
		It holds for $t=t_r$ where $\f(t_r,l_r|r)=0$ as line 1 of Algorithm \ref{Algo1} shows. Assume it holds for all $(t,l)$ with $t>t'$. Then we are to show it is correct for $t=t'$. By the induction hypothesis and Equation \ref{eqa}, the platform will always choose an available action $a$ with the highest $\mathcal{V}^{a}_s+\f(t_a,l_a|r).$ In line 5-7 and as visualized in Figure \ref{figthm1}, the platform will look in the order of $a_1,\ldots,a_j$ and pick the first location $a_i$ that is a destination of a request $r\in\cR_{t',l}$. Otherwise, the driver will be guided to drive idly to $d^*$. Let $P_i=\Pr[X_{l,a_i,t'}\geq 1|X_{l,a_k,t'}=0,\,\forall k<i]$, thus we have
		\begin{eqnarray*}
			&&\mathbb{E}_{\mathcal{R}_{t',l}}\left[v_{\pi}(t',l,\mathcal{R}_{t',l})\right]\\
			&=&\sum_{i=1}^jP_i\cdot(V_{l,a_i,t'}+\f(t'+\dist(l,a_i),a_i|r))\\
			&&\,+\Pr[X_{l,a_k,t'}=0,\,\forall k\leq j]\cdot \f(t'+\dist(l,d^*),d^*|r)\\
			&=&\f(t',l|r).
		\end{eqnarray*}
		The last equality follows from line 8-13, which concludes the proof.
		It follows $q_{\pi^*}(s,a)=\mathcal{V}_s^a+\f(t_a,l_a|r)$ as a corollary.
	\end{proof}
	
	\section{Algorithm for \UpdateDist}\label{AlgoUpdateDist}
	
	We show in Algorithm \ref{Algo5} the update of distribution. It calculates $p_w$'s following the intuition as we introduced (line 1-3) and then  the distribution is updated following Equation \eqref{eq1} (line 4-6). 
	Algorithm \ref{Algo4} shows the calculation of $p_w$'s, which follows a routine similar to calculating the CST value in Algorithm \ref{Algo1} 
	\newcommand{\GetProb}{{\sc GetProbability}}
	\begin{algorithm}[t]
		\caption{\GetProb$(l,t,x,h(\cdot)|r)$}\label{Algo4}
		\begin{algorithmic}[1]
			\If {$l=l_r \land t=t_r$}  \Return
			\EndIf
			\State Retrieve the CST values $\f(d,t+\dist(l,d)|r)$ for $d\in S(l,t|r)$
			\State $d^*\gets\arg\max_{d\in S(l,t|r)} \f(d,t+\dist(l,d)|r)$
			\State Denote $\{a_i\}$ the sequence of $d\in S(l,t|r)$ in decreasing order of $V_{l,d,t}+\f(d,t+\dist(l,d)|r)$
			\State  $j$ $\gets$ the largest index of $\{a_i\}$ such that $V_{l,a_j,t}+\f(a_j,t+\dist(l,a_j)|r)>\f(d^*,t+\dist(l,d^*)|r)$
			\State $p\gets 1$
			\For {$i=1$ to $j$}
			\State $w\gets (l,a_i,t)$
			\State $p_w\gets p_w+x\cdot p$
			\State \GetProb$(a_i,t+\dist(l,a_i),x\cdot p\cdot h(X_w\geq 1),h(\cdot)|r)$
			\State $p\gets p\cdot (1-h(X_w\geq 1))$
			\EndFor
			\State \GetProb$(d^*,t+\dist(l,d^*),x\cdot p,h(\cdot)|r)$
			
		\end{algorithmic}
	\end{algorithm}

	\begin{algorithm}[t]
		\caption{\UpdateDist$(h(\cdot),a,r)$}\label{Algo5}
		\begin{algorithmic}[1]
			\State Initialize $p_w\gets0$ for all $w\in\mathcal{W}$
			\State $(l_{a},t_{a})\gets$ the location-time pair action $a$ leads to
			\State \GetProb$(l_{a},t_{a},1,h(\cdot)|r)$
			\For{$w\in \cW$}
			\For{$i=1$ to $|\cD|$}
			\State $h(X_w\geq i)\gets (1-p_w)\cdot h(X_w\geq i)+p_w\cdot h(X_w\geq i+1)$
			\EndFor
			\EndFor
			\State \Return $h(\cdot)$
		\end{algorithmic}
	\end{algorithm}
	
	\section{Proof of Theorem \ref{thmworstcase}}\label{proofThm2}
	\begin{proof}
		Consider a graph of 4 vertices $A,B,C,D$. Let the weights of edges $\dist(B,D)=\dist(C,A)=\dist(A,D)=\mu$, $\dist(A,B)=\dist(D,C)=\mu-1$ and $\dist(B,C)=t$ ($1\leq t\leq\mu$), which represent the number of time steps required to travel along the edges as shown in Figure \ref{figthm2}.
		Suppose that $T=4\mu-2+t$ and there is only 1 vehicle. The vehicle starts work at $A$ at time $1$.
		
		Consider an instance on this graph and time horizon $[T]$ where we have 6  requests $r_1=(B,C,2\mu,t), r_2=(A,D,1,\mu), r_3=(D,C,\mu+1,\mu-1), r_4=(C,B,2\mu,t), r_5=(B,A,2\mu+t,\mu-1),r_6=(A,D,3\mu+t-1,\mu)$.
		
		The platform receives sufficiently many of $r_1$'s at the beginning and the algorithm will take one of $r_1$'s with a total revenue of $t$. This is because a deterministic algorithm should always accept the first feasible request, otherwise it would lead to a competitive ratio of $\infty$.
		
		After that, $r_2, r_3,\ldots, r_6 $ appear sequentially but the platform could accept none of them.
		
		In this case, the offline optimal solution would have taken all the requests except $r_1$ to fill up the entire time horizon with a total revenue of $4\mu-2+t$. Thus we have the ratio to be at least $\frac{4\mu-2+t}{t}|_{t=1}=4\mu-1$.

		\begin{figure}
			\begin{center}
				
				\includegraphics[width=5cm]{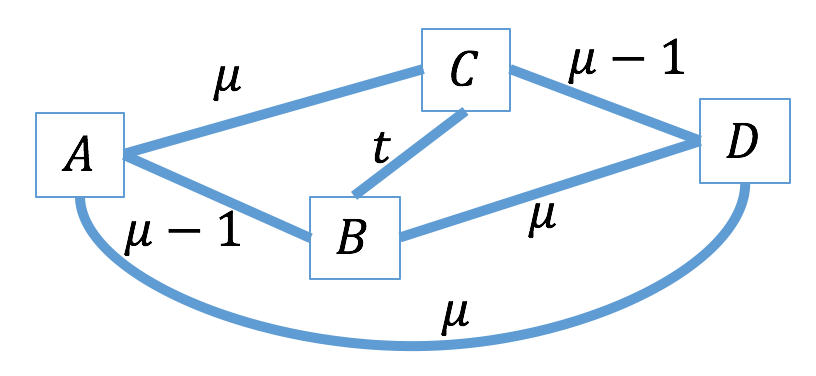}

				\caption{A Road Network Graph}  
				\label{figthm2}
			\end{center}
			
		\end{figure}
	\end{proof}
	
	\section{Proof of Theorem \ref{thmgreedy}}\label{proofThm3}
	\begin{proof}
		Let $R_{\ALG}$  be the set of requests the First-Fit algorithm accepts and  $R_{\OPT}$ be the one of the offline optimal solutions. For any request $r=(o_r,d_r,t_r,v_r)\in R_{\ALG}$, assume it is served by driver $c$. 
		Let $t_{r,1}=t_r$ and $t_{r,2}=t_r+\dist(o_r,d_r)$. Consider a time interval $I=[t_{r,1}-(2\mu-1),t_{r,2}+(2\mu-1)]$. 
		
		We claim that the travel time interval  of any request $r'$  that is incompatible with request $r$ and driver $c$, lies within $I$. 
		This is because if the time interval of a request does not lie entirely in $I$, then the end time of this request should be no later than $t_{r,1}-\mu$ (start time of  this request should be no earlier than $t_{r,2}+\mu$), thus it should be compatible with $r$.
		
		Thus, the total value of the requests in $R_{\OPT}$ incompatible with $r$ and driver $c$,  is at most $4\mu-2+t_{r,2}-t_{r,1}.$ Let $\OPT=\sum_{r\in\cR_{\OPT}}v_r$ and $\ALG=\sum_{r\in\cR_{\ALG}}v_r$, hence $\OPT\leq \sum_{r\in R_{\ALG}}(4\mu-2+t_{r,2}-t_{r_1}).$ As a result,
		\begin{eqnarray*}
			\frac{\OPT}{\ALG}&=&\frac{\OPT}{\sum_{r\in R_{\ALG}}t_{r,2}-t_{r,1}}\\
			&\leq& \frac{\sum_{r\in R_{\ALG}}4\mu-2+t_{r,2}-t_{r,1}}{\sum_{r\in R_{\ALG}}t_{r,2}-t_{r,1}}\\
			&=& \frac{|R_{\ALG}|(4\mu-2)}{\sum_{r\in R_{\ALG}}t_{r,2}-t_{r,1}}+1
			\\&\leq& \frac{|R_{\ALG}|(4\mu-2)}{|R_{\ALG}|}+1=4\mu-1.
		\end{eqnarray*}
		
	\end{proof}
	
	\section{Offline Algorithm for Stage 1 without On-Demand Requests}\label{FlowBased}
	The offline optimal solution of Stage 1 without on-demand requests can be obtained by solving a maximum cost network flow (MCNF).
	
	Given the set of available vehicles $\cD$ and all the scheduled requests $\cR$, we construct a network $G =(V,E)$. We construct two vertices $v_i$ and $v_i'$ for each vehicle $i$, one entry-vertex $v_{r,in}$ and one exit-vertex $v_{r,out}$ for each request $r$, 2 virtual vertices $S$ and $T$ as the global source and sink.
	
	We construct four types of edges in $E$. First, we construct edges from $S$ to $v_i$ and from $v_i'$ to $T$, each with flow $1$ and cost $0$. 
	Secondly, we construct edges from $v_{r,in}$ to $v_{r,out}$, with flow $1$ and cost $v_r$, which mean each request could be taken no more than once. Thirdly, we construct edges from $v_i$ to $v_{r,in}$ and from $v_{r,out}$ to $v_i'$, each with flow $1$ and cost $0$, which mean the first and last request the vehicle $i$ could possibly served. Lastly, we construct edges from $v_{r_i,out}$ to $v_{r_j,in}$ if the distance between region $i$ and region $j$ allows a vehicle to pick up request $j$ after serving request $i$.
	
	Then by applying any MCNF algorithm, we  could obtain the optimal solution.
	
	\section{The Greedy-KM and Enhanced-KM}\label{matchBaseline}
	Greedy-KM works as follows.
	Given the set of available vehicles $\cD$ and the state $(t_c,l_c,\cR_{t_c,l_c})$ of each vehicle, we construct a bipartite graph $G_B=(\cD,\bigcup_{c\in\cD}\cR_{t_c,l_c},E_B)$, where we have edges between $c\in\cD$ and $r\in\cR_{t_c,l_c}$ with weight $v_r$.
	Greedy-KM dispatches order by finding a weighted maximum matching on $G_B$. In implementation, we employ the Kuhn-Munkres (KM) algorithm \cite{munkres1957algorithms} to solve it. 
	
	In Enhanced-KM, the bipartite graph is constructed in the same way as Greedy-KM, except that the edges between $c\in\cD$ and $r\in\cR_{t_c,l_c}$ have weight $v_r+\f(t_c+\dist(o_r,d_r),d_r|\tilde{r_c})$, where $\tilde{r_c}$ is the next committed scheduled request of vehicle $c$.

	\section{Learning \& Planning Algorithm}\label{LPA}
	The LPA is an adaptation of the work of Xu et al.~\shortcite{xu2018large} to the hard constraints brought in by the scheduled requests.
	
    In their work, they regard consider the transportation as the MDP and construct a local-view MDP for each driver, with location-time pairs as the states. As for the state transition rules and rewards for each state, they are drawn from the historical data. Actions of drivers are to pick up on-demand requests nearby or to stay still. 
	For an action that lasts for $T'$ time steps with reward $R$,
	they apply a discount factor $\gamma$ and the final reward is given by
	\begin{eqnarray*}
		R_{\gamma}=\sum\limits_{t=0}^{T'-1}\gamma^t\frac{R}{T'}.
	\end{eqnarray*}
	At every time step, they obtain the value function $v'$ for all states and then dispatch orders via a matching approach. 
	The calculation of the value function is shown as Algorithm \ref{Algo6}. We are using the same notation in Algorithm \ref{Algo6} as Xu et al.~\shortcite{xu2018large}  did, which do not have the same meaning as those in our main text.
	
	We do the following to adapt their algorithms to our two-stage model. In Stage 1, we parse the scheduled requests and decide immediately for request $r$ by the comparison of $R_{\gamma}(r)+V'(d_r,t_r+\dist(o_r,d_r))$ and $V'(o_r,t_r)$, meaning that a request would be accepted if it could lead to an increment in the expected value. In Stage 2, we will use the same reward function as the edge weights for all the possible state transitions. It is worth noting that we will forbid the driver to pick up an order whose ending time is too late for next scheduled request.
	
	Moreover, Xu et al.~\shortcite{xu2018large} mentioned the updates in Algorithm \ref{Algo6} can be done iteratively with the planning. The time window in our algorithm, however, is only one day. Thus this iteration cannot help optimize the value function.
	
	\begin{algorithm}[!htbp]
		\caption{LPA}\label{Algo6}
		\begin{algorithmic}[1]
			\State Collect all the historical state transitions $\mathcal{T}=\{(s_i,a_i,r_i,s_i')\}$ from data; each state is composed of a time-location pair: $s_i = (t_i,location_i)$; each action is composed of the initial state and transited state: $a_i = (s_i, s_i')$; 
			\State Initialize $V'(s)$, $N'(s)$ as zeros for all possible states. 
			\For{$t = T - 1$ to $0$}
			\State Get the subset $D^{(t)}$ where $t_i=t$ in $s_i$.
			\For{each sample $(s_i,a_i,r_i,s_i') \in D^{(t)}$}
			\State $N'(s_i) \leftarrow N'(s_i) + 1$
			\State $V'(s_i) \leftarrow V'(s_i) + \frac{1}{N'(s_i)}[\gamma^{\Delta t(a_i)}V'(s_i')+R_{\gamma}(a_i)-V'(s_i)]$
			\EndFor
			\EndFor
			\State Return the value function $V'(s)$ for all states.
			
		\end{algorithmic}
	\end{algorithm}
	
	
	
	
	\section{Clustered Centers in the City}
\begin{figure}
		\begin{center}
			\includegraphics[width=6cm]{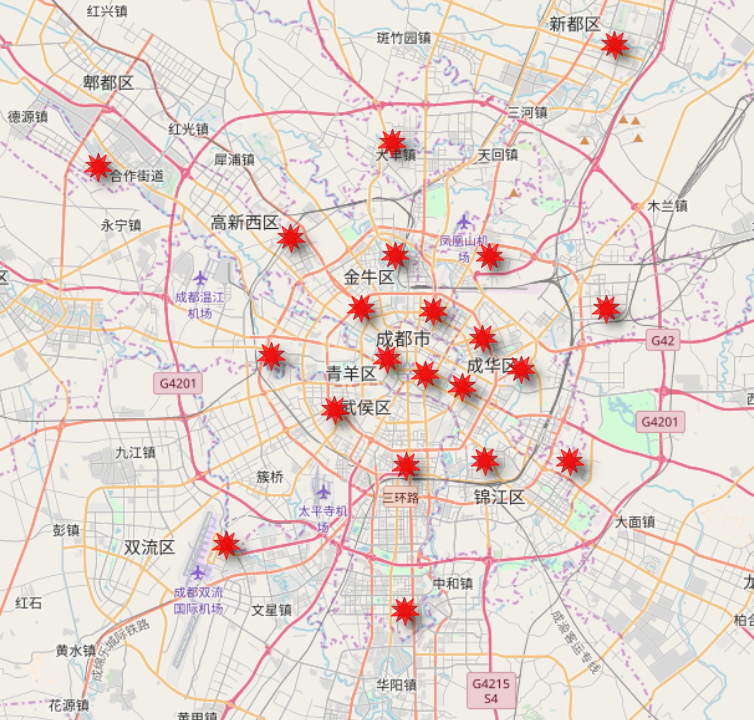}
			\caption{The Clustered Centers in a City in China derived from data.}
			\label{figmap}
		\end{center}
	\end{figure}

	In figure \ref{figmap}, we display the explicit real-world locations of 21 clustered centers in the Chinese city derived from the Didi data. 
	
	\section{Single-Vehicle Case Performance}\label{append:SV}
\begin{figure}[!h]
		\begin{center}
			\includegraphics[width=7cm]{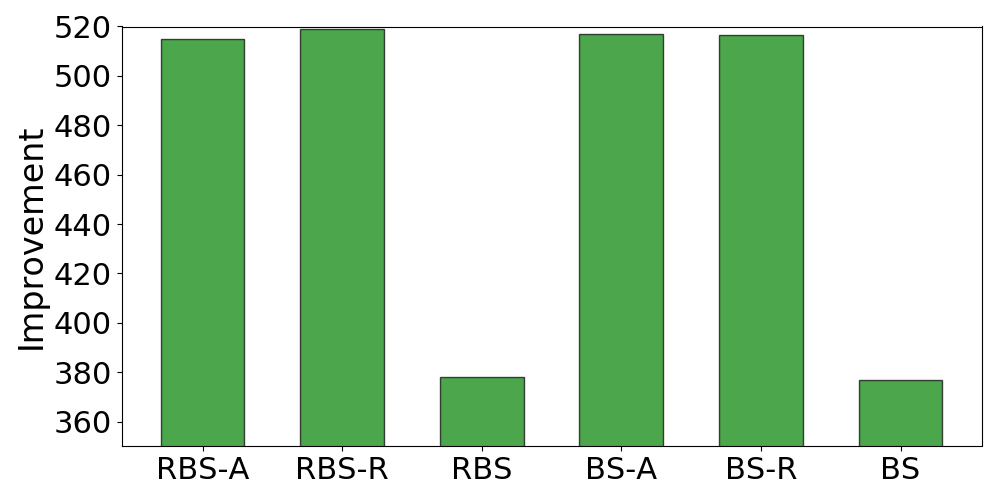}
			\caption{The Performance in Single-Vehicle Case.}
			\label{figsgv}
		\end{center}
	\end{figure}

	In Figure \ref{figsgv}, we demonstrate the single-vehicle performances among all \BestFit, \RandomBestFit,\BestFitA, \BestFitR, \RandomBestFitA, \RandomBestFitR, and label them as BS, RBS, BS-A, BS-R, RBS-A, RBS-R respectively in the figure.
	
	\section{CST-Value as the Vehicle Number Increases}\label{append:CSTVN}
\begin{figure}[!h]
		\begin{center}
			\includegraphics[width=7cm]{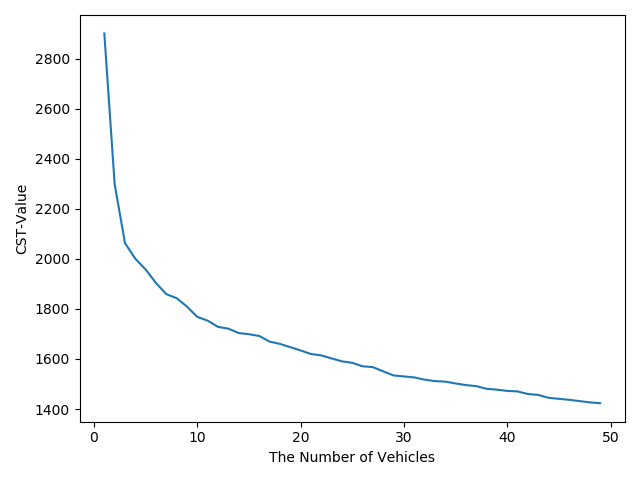}
			\caption{The Change of CST-function as Vehicles Increase.}
			\label{figcstvn}
		\end{center}
	\end{figure}

	Here we empirically demonstrate how the CST-value changes as the number of vehicles increases. We fix a tuple of $(t,l,r)$ and assume all vehicles start at $(t,l)$ with $r$ the next request to serve. Figure \ref{figcstvn} shows how $\f(t,l|r)$ changes and we can see that the decrease of $\f(t,l|r)$ is quick at first and later slows down.

\end{appendices}

\end{document}